\documentclass[runningheads]{llncs}
\usepackage[T1]{fontenc}

\usepackage{subcaption}

\usepackage{graphicx}%
\usepackage{multirow}%
\usepackage{amsmath,amssymb,amsfonts}%
\usepackage{amsthm}%
\usepackage{mathrsfs}%
\usepackage[title]{appendix}%
\usepackage{xcolor}%
\usepackage{textcomp}%
\usepackage{manyfoot}%
\usepackage{booktabs}%
\usepackage{algorithm}%
\usepackage{algorithmicx}%
\usepackage{algpseudocode}%
\usepackage{listings}%

\usepackage{graphicx,xfp}
\usepackage{multicol}
\usepackage{multirow}
\usepackage{color}
\usepackage{xcolor}
\usepackage{booktabs}
\usepackage{pifont}%
\newcommand{\xmark}{\ding{55}}%
\usepackage{textcomp}
\usepackage{sidecap}
\usepackage{graphicx,xfp}
\usepackage{amsthm}

\usepackage{caption}

\begin{document}

\title{Efficient Classification of SARS-CoV-2 Spike Sequences Using Federated Learning}









%
\titlerunning{FL-based SARS-CoV-2 Spike Sequences Classification}
%
\author{Prakash Chourasia\inst{1}
\and
Taslim Murad\inst{1}
\and
Zahra Tayebi\inst{1}
\and
Sarwan Ali\inst{1}
\and
Imdad Ullah Khan\inst{2}
\and
Murray Patterson\inst{1}
\\
\{pchourasia1, tmurad2, ztayebi1, sali85\}@student.gsu.edu, imdad.khan@lums.edu.pk, mpatterson30@gsu.edu
}
\authorrunning{P. Chourasia et al.}
%
\institute{Georgia State University, Atlanta GA 30303, USA 
\and
Lahore University of Management Sciences, Lahore Punjab 54792, Pakistan
}
\maketitle

\begin{abstract}
This paper presents a federated learning (FL) approach to train an AI model for SARS-Cov-2 variant classification. We analyze the SARS-CoV-2 spike sequences in a distributed way, without data sharing, to detect different variants of this rapidly mutating coronavirus. 
Our method maintains the confidentiality of local data (that could be stored in different locations) yet allows us to reliably detect and identify different known and unknown variants of the novel coronavirus SARS-CoV-2. 
Using the proposed approach, we achieve an overall accuracy of $93\%$ on the coronavirus variant identification task. 
We also provide details regarding how the proposed model follows the main laws of federated learning, such as Laws of data ownership, data privacy, model aggregation, and model heterogeneity.
Since the proposed model is distributed, it could scale on ``Big Data'' easily.  We plan to use this proof-of-concept to implement a privacy-preserving pandemic response strategy.
\end{abstract}


\section{Introduction}
The COVID-19 pandemic, caused by the SARS-CoV-2 coronavirus,
has impacted the entire
globe~\cite{majumder2021recent}.
It is responsible for almost $6$ million in deaths and $561$ million infected people as of July 2022 as reported by the World Health Organization
(WHO)~\cite{covid_stats_who}. This influence has drawn the attention of the research community to actively contribute their tools and techniques toward pandemic response strategies, such as the design and assessment of containment measures~\cite{kaimann2021containment,coccia2020impact}, image processing for  diagnosis~\cite{udugama2020diagnosing,panwar2020application}, optimal vaccine distribution~\cite{AHMAD2020Combinatorial,Tariq2017Scalable,ahmad2017spectral,lee2021performance}, computational tomography for genome
sequencing~\cite{udugama2020diagnosing}, etc.

Moreover, to comprehend the diversity and dynamics of the virus, its genome sequences are analyzed by using phylogenetic methods~\cite{hadfield2018a,minh_2020_iqtree}. These methods can help in variant identification, however, they are not scalable~\cite{hadfield2018a,minh_2020_iqtree}. Due to the presence of large amounts of publicly available biological sequence data on databases such as GISAID~\cite{gisaid_website_url}, it is desirable to design a scalable analytical model to get a deeper understanding of the virus.

\begin{figure}[h!]
  \centering
  \includegraphics[width=0.8\linewidth] {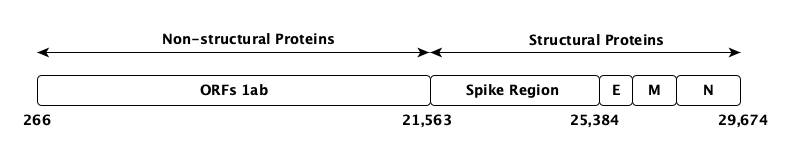}
  \caption{The SARS-CoV-2 genome is roughly 30kb in length, encoding
    structural and non-structural proteins.  The spike region is composed of 3821 base pairs.}
    \label{fig_spikeprot}
\end{figure}

Furthermore, the detailed SARS-CoV-2 genome structure is illustrated in Figure~\ref{fig_spikeprot}. It consists of many sub-parts including the spike region, which is essential because the virus attaches to the host cell through this region. It also contains many of the mutations of the SARS-CoV-2 virus, which can result in creating different variants of this virus. Therefore, rather than using the full genome sequence of the virus, the spike sequence alone provides sufficient information to reliably analyze this virus. Recently, classification and clustering approaches are proposed to analyze the SARS-CoV-2 virus using only spike sequences, like host classification~\cite{kuzmin2020machine,ali2022pwm2vec}, variant classification~\cite{ali2021k,tayebi2021robust,ali2021spike2vec}, etc. These methods first generate numerical embeddings of the sequences and then employ either vector-space or kernel-based classifiers.

Traditionally, the training of a machine learning (ML) model happens in a centralized way with all of the data stored on or is
available to be sampled from a single
server~\cite{kairouz2021advances}.
However, privacy and data security
concerns discourage disparate entities (e.g., healthcare governing
bodies in different countries) from sharing the data. The under-reporting
of COVID-19 statistics and other related data has already been
observed in various regions~\cite{kisa2020under,xu2020covid}, due to
political or other reasons. Even in cases where there are no
ethical, regulatory, or legal issues in data sharing, healthcare
bodies are known to prefer models validated on their
data~\cite{buch2021development}. Moreover, the local context is
already lost in a model trained on global data. On the other hand,
models trained on ``limited local'' data tend to overfit and do not
generalize.

\begin{figure}[h!]
  \centering
  \includegraphics[scale=0.20]{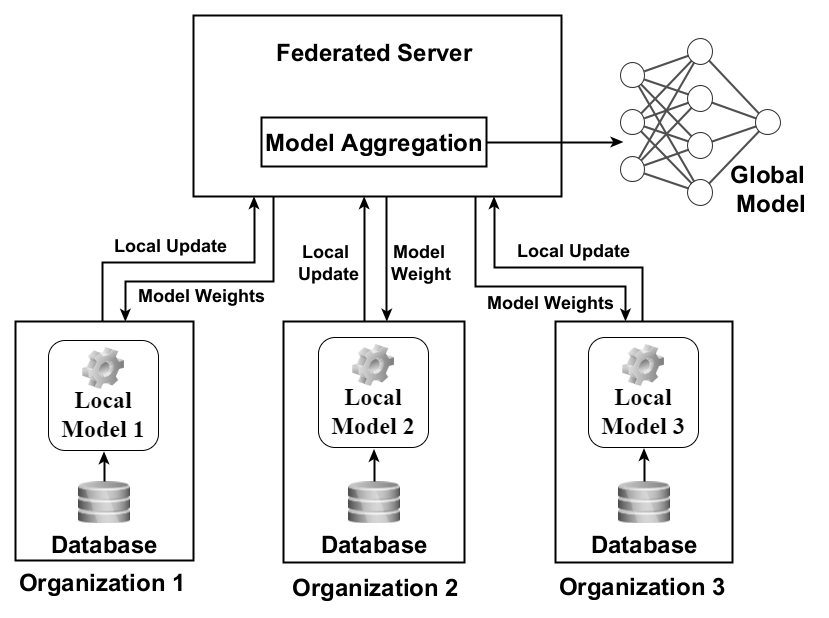}
  \caption{The federated learning approach for a learning task using
    private data from three separate organizations.}
  \label{fig_federated_learning}
\end{figure}

Federated learning (FL), an AI paradigm, offers a more pragmatic and
proven approach to dealing with many facets of data-sharing
challenge. FL~\cite{mcmahan2017communication} enables collaborative
model learning over data from several (decentralized) places without
any data relocation. In FL, as shown in
Figure~\ref{fig_federated_learning}, first, (many) local models are
trained using the private data at each location. A \emph{global
  model} is then trained using \emph{federated learning}. The
global model is kept on a central server called a \emph{federated
  server}. Model parameters from the local models are pushed onto the
federated server, aggregating them using an
\emph{aggregation function}. FL preserves data privacy, overcomes
data ownership barriers, and yields generalized models.

The concept of federated learning has been used in many different
areas~\cite{aledhari2020federated,shaheen2022applications,chourasia2023empowering}, including
mobile apps, internet-of-things (IoT), transportation, bioinformatics and defense. Due to its
applicability and the numerous trials that have previously been done,
 it is quite dependable. Recently, FL has been suggested for
inter-institutional healthcare research considering its core principle
where only model weights or gradients are sent between client sites
and the federated server, easing privacy concerns about data
governance for FL~\cite{dayan2021federated}.

In this paper, we build a small prototype of federated learning (FL)
model using a set of spike sequences for coronavirus variant
classification. We compare the performance of our proposed FL-based
approach on spike sequence data versus expensive baseline methods. In
addition, we compare our proposed solution with other traditional
state-of-the-art (SOTA) approaches, which involve a centralized model
training approach using different embedding methods to address the
classification problem.

We envision the use of an FL-based solution as a solution for authorities and governments to facilitate different privacy and simultaneously extract the knowledge from these large public (global) datasets (repositories such as GISAID) along with private (local) datasets from other countries (private dataset) for a customized model catered to solving public health issues and designing policies in a specific context (e.g.,
state, country, geographical region). Here, we propose a federated learning-based approach to efficiently utilize a publicly available data set (i.e. GISAID), and a mechanism to extract helpful information from the
private data of others while facilitating the differential privacy of
contributors to the problem of classifying variants of the SARS-CoV-2
virus. 
For this purpose, we extracted $9000$ spike protein sequences from GISAID along with their lineage information to perform multi-class classification. Our dataset comprised $9$ unique lineages.
Our scalable model provides a framework for solving similar problems
in other domains. Moreover, we show that using the spike protein
instead of the whole genomic sequence can give the same or better
accuracy, thus reducing computational complexity significantly.

Our contributions are as follows:

\begin{enumerate}
    \item For coronavirus spike sequence classification, we provide federated learning (FL) based models, which are scalable and can be applied in a distributed fashion with less computational overhead.
    \item Using the proposed FL model in a distributed manner allows us to maintain data privacy by only sending outputs (differential privacy) from the local small models to the global model (secure multi-party computation).
    \item We compare FL-based models with different state-of-the-art
    (SOTA) embedding techniques and show that the proposed model
    outperforms SOTA methods in terms of predictive accuracy.
    \item We demonstrate that the underlying machine learning classifiers can achieve high predictive performance with a fraction of the information (spike sequences rather than full-length genome sequences).
\end{enumerate}

The rest of the paper is organized as follows:
Section~\ref{sec_related_work} contains the related work. Our proposed federated learning model is explained
in detail in Section~\ref{sec_proposed_approach}.
Section~\ref{sec_experimental_setup} provides the details on the
dataset and experimental setup.  Results are given in
Section~\ref{sec_results_discussion}, and we conclude the paper in
Section~\ref{sec_conclusion}.

\section{Related Work}\label{sec_related_work}
There are several approaches to convert biological sequences into machine learning-compatible inputs for classification and clustering, like $k$-mers-based methods~\cite{wood-2014-kraken,ali2021k,ali2021effective,solis-2018-hiv}. 
Similarly, a position weight matrix (PWM) based classification approach is proposed in~\cite{ali2022pwm2vec}, which generates a fixed-length representation of spike sequences based on weights of $k$-mers computed using PWM.

Although the methods discussed above show higher predictive performance, they do not consider the privacy of data. To ensure the privacy of the information, a novel technique called federated learning (FL) has caught the attention of researchers. In~\cite{nasser2022lightweight}, authors use the data gathered by individual user entities/equipment utilizing ambient sensors and wearable devices to propose a lightweight FL model that may be used to privately and collectively learn medical symptoms (like COVID-19). 
Moreover, Many FL-based methods for image classification are put forward, like the authors in~\cite{li2021model} proposed MOON framework to deal with the heterogeneity of data distribution among local parties in FL. 
In another work~\cite{jimenez2021memory}, early breast cancer prediction is made by a memory-aware curriculum federated learning-based model using mammography images.  
The system given in~\cite{li2020multi} is performing neuroimage analysis by following an FL-based strategy. In~\cite{zhang2021dynamic} authors used FL for COVID detection using x-ray images. Using data from 20 institutions throughout the world, the authors in~\cite{dayan2021federated} proposed a model called EXAM (electronic medical record (EMR) chest X-ray AI model). However, the model uses inputs of vital signs, laboratory data, and chest X-rays to forecast the future oxygen requirements of symptomatic COVID-19 patients. It is heterogeneous but is clinical and image data. Unlike these image-based approaches, our proposed method directly works on the sequence data. Although there are studies related to medical federated learning, specifically for Oncology and Cancer Research~\cite{chowdhury2021review} along with biases in the genomic data collection to perform federated learning~\cite{boscarino2022federated}, these studies do not present an end-to-end federated learning-based pipeline to perform privacy aware spike sequence classification.

\section{Proposed Approach}\label{sec_proposed_approach}
In this section, we describe the proposed FL-based approach for the classification of coronavirus variants from spike protein sequences. We explain in detail the overall architecture of the proposed model.

\subsection{Architecture}
The architecture consists of two types of components: 1) client models (local) and 2) Federated Learning models (global). The approach is based on a decentralized data approach that involves dividing the dataset into different smaller parts and processing each part separately. The client model is composed of three parts of the dataset to train the models locally. These trained local models are pushed to a central (global) neural network (NN) based model. Only the weights, biases, and other parameters are provided to the global NN. To further reduce the size of the global model, it may undergo pruning (removing the less important parameters).
The NN model gets all the locally trained models and averages them out, effectively creating a new global model (Federated Learning model). The Federated Learning model coordinates the federated learning process and uses a fourth part of the dataset to train the global model. Each step is explained in more detail below:

\subsection*{Step 1: Feature Vector Generation}
A fixed-length numerical feature vector called One Hot Encoding (OHE) is proposed in~\cite{ali2021k,kuzmin2020machine}. It generates a binary ($0-1$) vector based on the character's position in the sequence given alphabet $\Sigma$, where $\Sigma$ is ``\textit{ACDEFGHIKLMNPQRSTVWXY}'', the unique characters in each sequence. The $0-1$ vectors for all characters are concatenated to make a single vector for a given sequence. For a given sequence $i$ of length $l$, the dimension of OHE based vector i.e., $\phi_i$ can be denoted by  $\phi_i = \vert \Sigma \vert \times l$.

\subsection*{Step 2: Federated Learning Approach}
After generating the numerical vectors $\phi$ for SARS-CoV-2 spike sequences, we use these feature vectors as input for our federated learning-based model.  We divide the dataset into training ($\phi_{tr}$) and testing ($\phi_{ts}$). The training dataset  $\phi_{tr}$ is further divided into four equal parts ($\phi_{tr1}, \phi_{tr2}, \phi_{tr3}, \phi_{tr4}$). 
Our final Federated Learning-based model is comprised of a local and a global model, which work together to classify the spike sequences.

In the current architecture, we divide the data randomly into different models with equal proportions. In a real-world scenario, the data distribution among local models could be very different as some models can be trained on more data compared to other models. However, since we are using simple machine-learning classifiers in the local models, and since they are not as data-hungry as typical neural networks, they could generalize easily to different distributions of data in different local models.

\subsubsection*{Local models}
We initialize $3$ individual classification (local) models (using classifiers such as XGB, Logistic Regression (LR), and Random Forest (RF)) and train them using three parts of the data ($\phi_{tr1}, \phi_{tr2}, \phi_{tr3}$). After training the ``local model", 
these models are used to create a new aggregated model (global). 

\subsubsection*{Global model}
Our global model consists of a neural network architecture, which takes $\lambda_{1}$, $\lambda_{2}$, and $\lambda_{3}$ as input where $\lambda_{1}$, $\lambda_{2}$, and $\lambda_{3}$ are the outputs from local trained models for the dataset $\phi_{tr4}$, thus training the neural network using $\phi_{tr4}$. It is important to point out that only the weights, biases, and other parameters are transferred 
to a new global model (from the local models). In the global model, none of the data from the three parts of the dataset ($\phi_{tr1}, \phi_{tr2}, \phi_{tr3}$) is used, which is the core concept of federated learning. Using the fourth part of the training data ($\phi_{tr4}$) we get output $\lambda_{1}$, $\lambda_{2}$, and $\lambda_{3}$ from respective trained classification models (local) for each data sample. This output of dimension $9\times 3 = 27$ (probability for $9$ class labels from $3$ models) is supplied to the neural network as input to train the Neural Network in the (global) model. We get our final trained ensemble model after this step.
Figure~\ref{fig_fl_NN} shows the precise architecture of the deep learning (DL) model, which is employed as the global model.
The number of neurons in the input layer is $27$ (weights from $3$ local model for $9$ class labels). The output layer, which has $9$ neurons, represents the nine classes we predict. 
The neural network has two hidden layers with $25$ and $15$ neurons, respectively. Each hidden layer has a ReLu activation function, while the final classification layer uses a Softmax function to handle our multi-class classification problem. Furthermore, we use the ADAM optimizer with $16$ batch size and $100$ training epochs as hyperparameters.
The number of parameters is listed in Table~\ref{tbl_nn_params}, the number of trainable parameters for hidden layer 1 is $700$, hidden layer 2 is $390$, and the output layer is $144$. In total, the global model uses $1254$ trainable parameters.

\begin{table}[h!]
\begin{minipage}[b]{0.45\linewidth}
    \centering
    \includegraphics[scale=0.18]{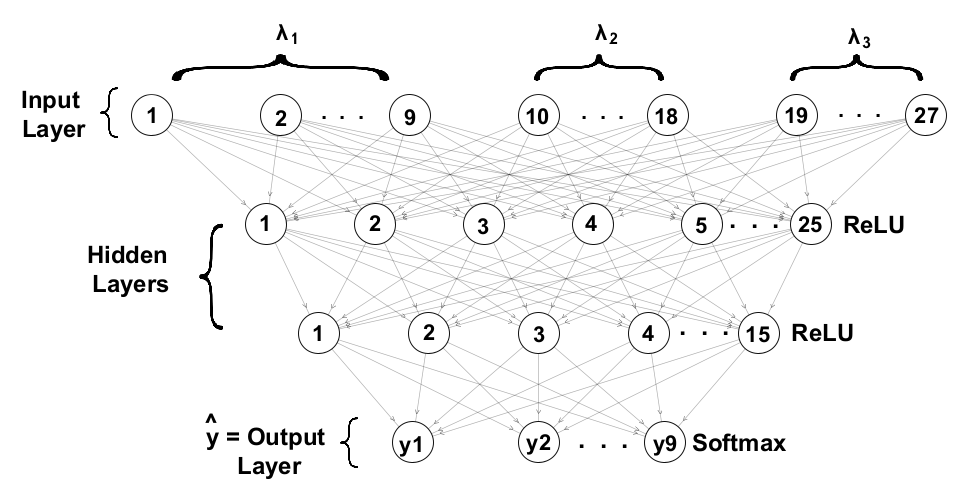}
    \captionof{figure}{Federated learning - Neural network.}
    \label{fig_fl_NN}
\end{minipage}
\hfill
\begin{minipage}[b]{0.50\linewidth}
\centering
    \resizebox{0.87\textwidth}{!}{
    \begin{tabular}{p{2.5cm}ccp{1.5cm}}
        \toprule
        Layer (type) & Input/Output Shape  & Trainable Parameters \\
        \midrule \midrule
        Input Layer & \begin{tabular}{c}Input : (None, 27) \\ Output : (None, 27)\end{tabular} & - \\
        \midrule
        Hidden Layer 1 & \begin{tabular}{c}Input : (None, 27) \\ Output : (None, 25)\end{tabular} & 700 \\
        \midrule
        Hidden Layer 2 & \begin{tabular}{c}Input : (None, 25) \\ Output : (None, 15)\end{tabular}  & 390 \\
        \midrule
        Output Layer & \begin{tabular}{c}Input : (None, 15) \\ Output : (None, 9)\end{tabular}  & 144 \\
        \midrule
        \midrule
        Total & \_  & 1254 \\
        \bottomrule
    \end{tabular}
    }
  \caption{Detail regarding the parameters in different layers of the Neural Network.}
  \label{tbl_nn_params}
\end{minipage}\hfill
\end{table}

\subsubsection*{Testing the ensemble model}
Finally, using the ensemble-trained global model, we predict for the test dataset $\phi_{ts}$ for the final predictions and evaluate our proposed model using different evaluation metrics. 

\subsection*{Workflow for proposed approach}
\begin{figure*}[h!]
\centering
  \includegraphics[width=0.95\linewidth] {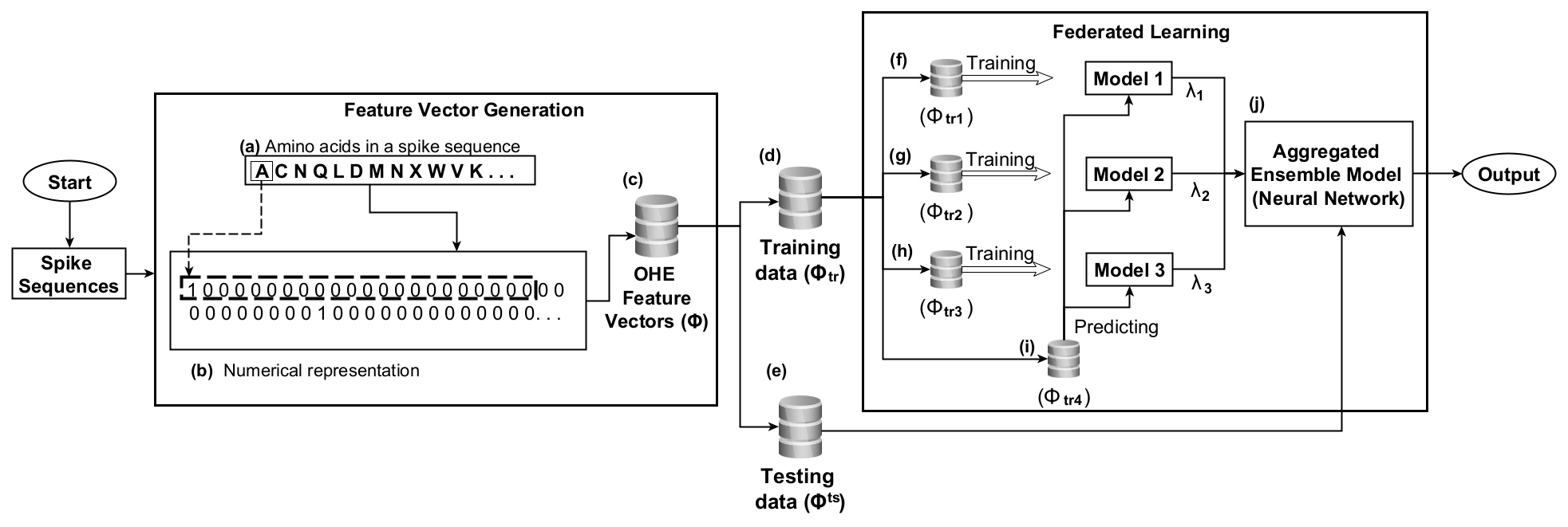}
  \caption{Flowchart of Federated Learning approach.
  }
  \label{fig_federated_learning_flowchart}
\end{figure*}

Figure~\ref{fig_federated_learning_flowchart} shows the complete workflow for our proposed approach. The left box shows the feature vector ($\phi$) generation process where we used One Hot Encoding to generate the numerical representation (feature vectors) from the spike sequences. Each amino acid in the spike sequence, as shown in Figure~\ref{fig_federated_learning_flowchart} (a), is encoded into numerical representation by placing $1$ at the position of a character. For example, for amino acid ``A" we place $1$ at the first position in the respective numerical representation as shown in (b). Afterward, we divide the feature vector dataset into training $\phi_{tr}$ and testing  $\phi_{ts}$. Box 2 on the right side of  Figure~\ref{fig_federated_learning_flowchart} shows our federated learning-based approach. We divide the training dataset into $4$ equal parts ($\phi_{tr1}, \phi_{tr2}, \phi_{tr3} \text{ and } \phi_{tr4}$) and use $3$ of these for training the ``local models" (e.g. random forest) as shown in Figure~\ref{fig_federated_learning_flowchart} (f-h). 
After training, these models are aggregated and assembled to create a new global model Figure~\ref{fig_federated_learning_flowchart} (j). The weights uploaded by each node (local model) for the training dataset $\phi_{tr4}$ are received on the server side as input. They are used to train the global neural network model. In the end, we use the testing dataset ($\phi_{ts}$) to predict and evaluate the model. 

The pseudo-code of our proposed approach is shown in Algorithm~\ref{algo_fedrated}. The given spike sequence-based data is converted to numerical vectors by employing a one-hot encoding technique. The resultant vectors are aligned following the trailing zero padding method. Then we split the aligned vectors into training and test sets.
The training set is further divided into four exclusive training sets, among which the three sets are used individually to train three local models respectively. We feed the fourth training set to the local models to obtain their respective weights.
Furthermore, we combine all the extracted weights and pass them to the global model as input. After the training, we employ the test dataset to get the predictions from the global model. These predictions can provide insight into the global model's performance.

\begin{algorithm}[h!]

	\caption{Ensemble Model Workflow.}
\label{algo_fedrated}
	\begin{algorithmic}[1]
 \scriptsize
	\State \textbf{Input:} Sequence data $S$
	\State \textbf{Output:} Sequences Variant Predictions $V$
	
	\State $\phi$ = OHE (S) \Comment{$ \text{get one-hot encodings of S }$}
	
	\State $\phi_{tr}$, $\phi_{ts}$ = SplitDataTrainTest ($\phi$)  \Comment{$ \text{70-30\% split}$} 
	
	\State $\phi_{tr1}$, $\phi_{tr2}$, $\phi_{tr3}$, $\phi_{tr4}$ = SplitTrainingData ($\phi_{tr}$ )   
\newline\Comment{$ \text{split training data into 4 sets }$} 
	
	\State $model_1$ = Train ($\phi_{tr1}$)  \newline\Comment{$ \text{train local $model_1$ with $\phi_{tr1}$ training set}$} 
	\State $\lambda_1$ = $model_1$($\phi_{tr4}$)  
	
	\State $model_2$= Train ($\phi_{tr2}$)  \newline\Comment{$ \text{train local $model_2$ with $\phi_{tr2}$ training set}$} 
	\State $\lambda_2$ = $model_2$($\phi_{tr4}$) 
	
	\State $model_3$ = Train ($\phi_{tr3}$)  \newline\Comment{$ \text{train local $model_3$ with $\phi_{tr3}$ training set}$}
	\State $\lambda_3$ =  $model_3$( $\phi_{tr4}$)   
	
	\State $model_{g}$ = Train ($\lambda_1$ + $\lambda_2$ + $\lambda_3$)  \newline\Comment{$ \text{pass $\lambda_1$ + $\lambda_2$ + $\lambda_3$ as input to global $model_{g}$}$}
	
	\State $V$ = $model_g$($\phi_{ts}$)  \Comment{$ \text{$model_g$ output V for $\phi_{ts}$ }$}

    \State return($V$ )
	\end{algorithmic}
\end{algorithm}

\section{Experimental Setup}\label{sec_experimental_setup}
In this section, we detail the spike sequence dataset used for experimentation. Followed by the details of the baseline models. In the end, we talk about the evaluation metrics used to test the performance of the models.
All experiments are conducted using an Intel(R) Core i5 system @ $2.10$GHz having Windows 10 $64$ bit OS with 32 GB memory. 
For the classification algorithms, we use $70\%$ of the data for training and $30\%$ for testing. The data is split randomly and experiments are repeated $5$ times to report average results. For hyperparameter tuning, we used $10\%$ data from the training set as a validation set.

\subsection{Dataset Statistics}
We extract the spike sequence data from GISAID~\footnote{\url{https://www.gisaid.org/}}.
It is a popular database in the bioinformatics domain that provides free open access to a large amount of sequence data related to the SARS-CoV-2 virus and Influenza virus. Moreover, GISAID facilitates genomic epidemiology and real-time surveillance to monitor the emergence of new COVID-19 viral strains across the planet.
The extracted data contains $9$ coronavirus variants within $9000$ total sequences ($1000$ sequences for each variant) that are selected randomly. Detailed statistics of the dataset can be seen in Table~\ref{tbl_variant_information}. The variant information is used as class labels for classification.
Every sequence is associated with a lineage or variant. The variant is generated by certain mutations in the spike protein region. For example, the epsilon variant is created when the mutations S13I, W152C, and L452R happen in the spike region, where S13I means the amino acid S at position 13 is replaced by amino acid I. We use these sequence-based datasets to predict the corresponding variant names. 

\begin{remark}
Note that the spike sequences in our data are not of the same length. The average, minimum, and maximum length of sequences (in the whole data) is $1263.16$, $9$, and $1277$, respectively. We use data padding in one-hot encoding to get a fixed-length representation. The sequence length statistics for individual variants are given in Table~\ref{tbl_variant_information}.
\end{remark}

\subsection{Baseline Model}
We use the following models from the literature as baselines for the comparison of results with the proposed federated learning model.

\paragraph{Spike2Vec~\cite{ali2021spike2vec}}
Spike2Vec is a method to convert bio-sequences into numerical form for enabling ML-based classification of the sequences. A sequence of length $N$ will have $N - k + 1$ $k$-mers. For our experiments, we used $k=3$.

\paragraph{WDGRL: }
A neural network (NN) based method that takes the one-hot representation of biological sequence as input and designs an NN-based embedding method by minimizing loss~\cite{shen2018wasserstein}.

\paragraph{PWM2Vec~\cite{ali2022pwm2vec}:} 
Using the idea of the position-weight matrix (PWM), this technique is intended to generate fixed-length numerical embeddings. It starts by creating a $\vert \Sigma \vert \times k$ dimensional PWM matrix from a protein sequence, which comprises the count of each amino acid inside $k$-mers of the sequence. Each $k$-mer is given a numerical weight based on the counts. The final representation is generated by concatenating all of the weights.

\paragraph{String Kernel: }
Kernel Matrix-based method which designs $n \times n$ kernel matrix that can be used with kernel classifiers or with kernel PCA~\cite{hoffmann2007kernel} to get feature vector based on principal components~\cite{farhan2017efficient,ali2022efficient,validationSetApproach}.

\paragraph{ProteinBert: }
It is a pre-trained Transformer, a protein sequence model to classify the given biological sequence using Transformer/Bert~\cite{BrandesProteinBERT2022}.

A summary of the comparison of different baseline models and the proposed federated learning-based approach is also shown in Table~\ref{tbl_baseline_andproposed_summary}.

\begin{table}[h!]
\begin{minipage}[b]{0.45\linewidth}
  \centering
  \resizebox{0.99\textwidth}{!}{
  \begin{tabular}{@{\extracolsep{4pt}}p{1.5cm}p{3cm}p{1.1cm}p{1.5cm} p{1.6cm}ccc}
    \toprule
    & & & & & \multicolumn{3}{c}{Sequence Length} \\
    \cmidrule{6-8}
    \multirow{2}{*}{Lineage} & \multirow{2}{3cm}{Region of First Time Detection} & \multirow{2}{1.1cm}{Variant Name} &
    \multirow{2}{1.8cm}{No. Mut. S/Gen.} &  No. of sequences & \multirow{2}{*}{Min.} & \multirow{2}{*}{Max.} & \multirow{2}{*}{Avg.} \\
    \midrule \midrule	
    B.1.351   & South Africa~\cite{galloway2021emergence} & Beta & 9/21 & \hskip.1in 1000 & 9  &  1274  &  1260.46 \\
    B.1.427   & California~\cite{zhang2021emergence}  & Epsilon  &  3/5 & \hskip.1in 1000 & 100  &  1274  &  1272.18  \\
    B.1.429   &  California~\cite{who_website}  & Epsilon  & 3/5  & \hskip.1in 1000 & 100  &  1277  &  1271.93 \\
    B.1.525   & UK and Nigeria~\cite{who_website} & Eta & 8/16 & \hskip.1in 1000 & 32  &  1273  &  1257.19 \\
    B.1.526   & New York~\cite{west2021detection}  &  Iota & 6/16 & \hskip.1in 1000 &  9  &  1273  &  1266.62 \\
    B.1.617.2  & India~\cite{yadav2021neutralization}  &  Delta &  8/17  & \hskip.1in 1000 & 99  &  1273  &  1265.12 \\
    B.1.621   & Colombia~\cite{who_website} & Mu & 9/21 & \hskip.1in 1000 & 9  &  1275  &  1255.93 \\
    C.37  & Peru~\cite{who_website} & Lambda & 8/21 & \hskip.1in 1000 & 86  &  1273  &  1248.55 \\
    P.1   & Brazil~\cite{naveca2021phylogenetic} &  Gamma &  10/21  & \hskip.1in 1000 & 99  &  1274  &  1270.45 \\
    \midrule
    Total & -  & -  & -  & 9000 & - & - & - \\
    \bottomrule
  \end{tabular}
  }
  \caption{Statistics for $9$ lineages from the SARS-CoV-2 dataset. }
  \label{tbl_variant_information}
  \end{minipage}
  \hfill
\begin{minipage}[b]{0.50\linewidth}
\centering
    \resizebox{0.85\textwidth}{!}{
    \begin{tabular}{p{2.2cm}p{1.5cm}p{1.2cm}p{2.2cm}p{1.2cm}p{1.2cm}} 
    \toprule
    Embedding & Alignment Free & Privacy & Low Communication Cost & Space Efficient & Runtime Efficient \\
    \midrule	\midrule	
    Spike2Vec &  \hfil $\checkmark$ &  \hfil \xmark &  \hfil \xmark &  \hfil $\checkmark$ &  \hfil \xmark  \\
     \cmidrule{2-6}
     WDGRL &  \hfil \xmark &  \hfil \xmark &  \hfil \xmark &  \hfil $\checkmark$ &  \hfil \xmark  \\
     \cmidrule{2-6}
     PWM2Vec &  \hfil \xmark &  \hfil \xmark &  \hfil \xmark &  \hfil $\checkmark$ &  \hfil $\checkmark$  \\
     \cmidrule{2-6}
    String Kernel &  \hfil $\checkmark$ &  \hfil \xmark &  \hfil \xmark &  \hfil \xmark &  \hfil \xmark  \\
     \cmidrule{2-6}
    ProteinBert &  \hfil $\checkmark$ &  \hfil \xmark &  \hfil \xmark &  \hfil \xmark &  \hfil \xmark   \\
     \cmidrule{2-6}
   \multirow{2}{2.5cm}{Federated Learning (ours)} & \hfil $\checkmark$ & \hfil $\checkmark$ & \hfil $\checkmark$ & \hfil $\checkmark$ & \hfil $\checkmark$   \\  
   & \\
    \bottomrule
  \end{tabular}
  }
  \caption{Baseline and Proposed Methods advantages and disadvantages.}
  \label{tbl_baseline_andproposed_summary}
\end{minipage}\hfill
\end{table}

\subsection{Machine Learning Classifiers}
For the classification task on state-of-the-art methods, we use Support Vector Machine (SVM), Naive Bayes (NB), Multi-Layer Perceptron (MLP), K Nearest Neighbors (KNN) $K=5$, Random Forest (RF), Logistic Regression (LR), and Decision Tree (DT). 

For the FL, we use eXtreme Gradient Boosting (XGB), LR, and RF classifiers to train the local models. XGB is a boosting algorithm based on the gradient-boosted decision trees approach. It applies a better regularization technique to reduce over-fitting.
We select important features from the training dataset using a meta-transformer approach. This approach involves selecting features based on importance weights and is used for feature selection (dimensionality reduction). The goal of dimensionality reduction is to either improve the accuracy scores of the estimators or to boost the model's performance on high-dimensional datasets, hence avoiding the curse of dimensionality.

\subsection{Evaluation Metrics}
We use
average accuracy, precision, recall, weighted $F_1$, macro $F_1$, and ROC-AUC (one-vs-rest) metrics to evaluate the performance of classification algorithms. We also report the training runtime for the classifiers. Note that for the federated learning-based model, the reported runtime is for the whole end-to-end model.

\subsection{Data Visualization}
\begin{figure}[h!]
  \centering
  \begin{subfigure}{.25\textwidth}
  \centering
  \includegraphics[scale=0.09]{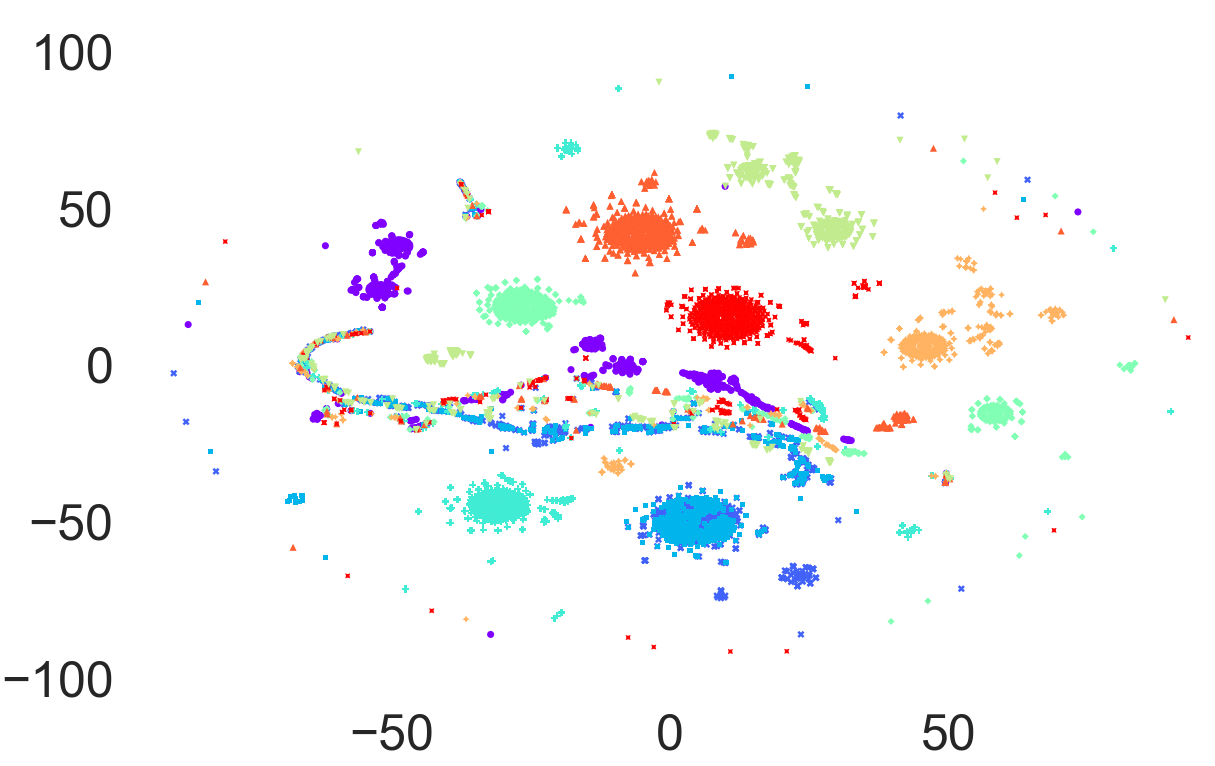}%
  \caption{Spike2Vec}
  \end{subfigure}%
  \begin{subfigure}{.25\textwidth}
  \centering
  \includegraphics[scale=0.09]{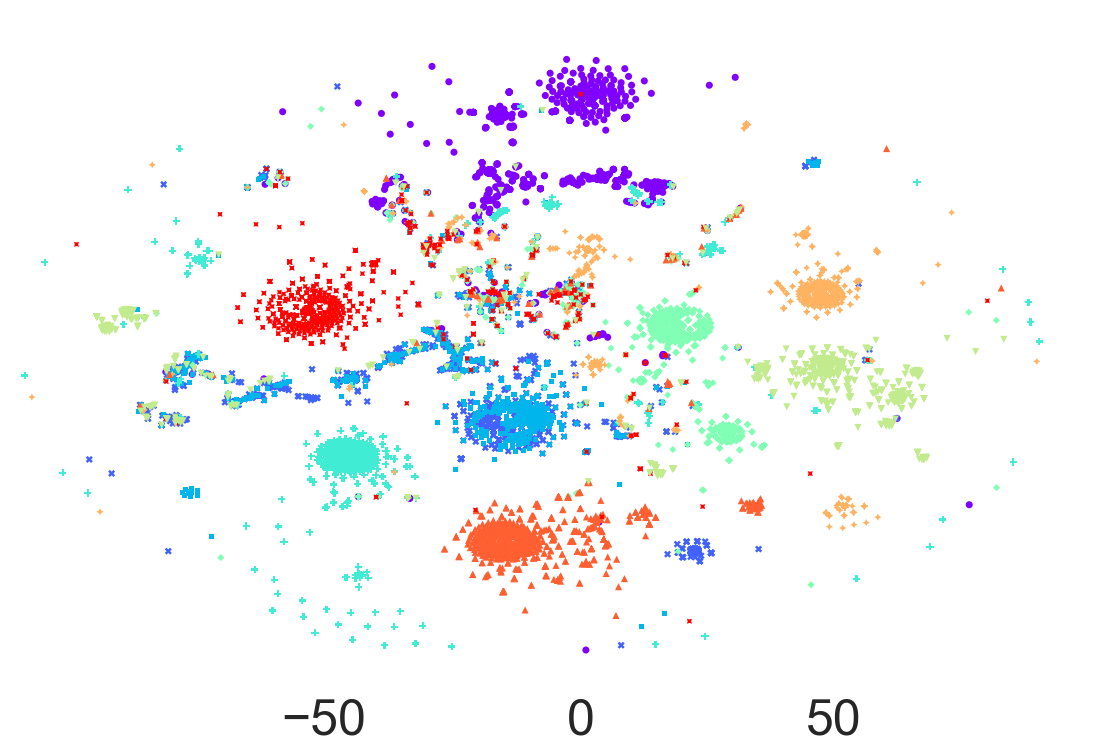}%
  \caption{PWM2Vec}
  \end{subfigure}%
  \begin{subfigure}{.25\textwidth}
  \centering
  \includegraphics[scale=0.09]{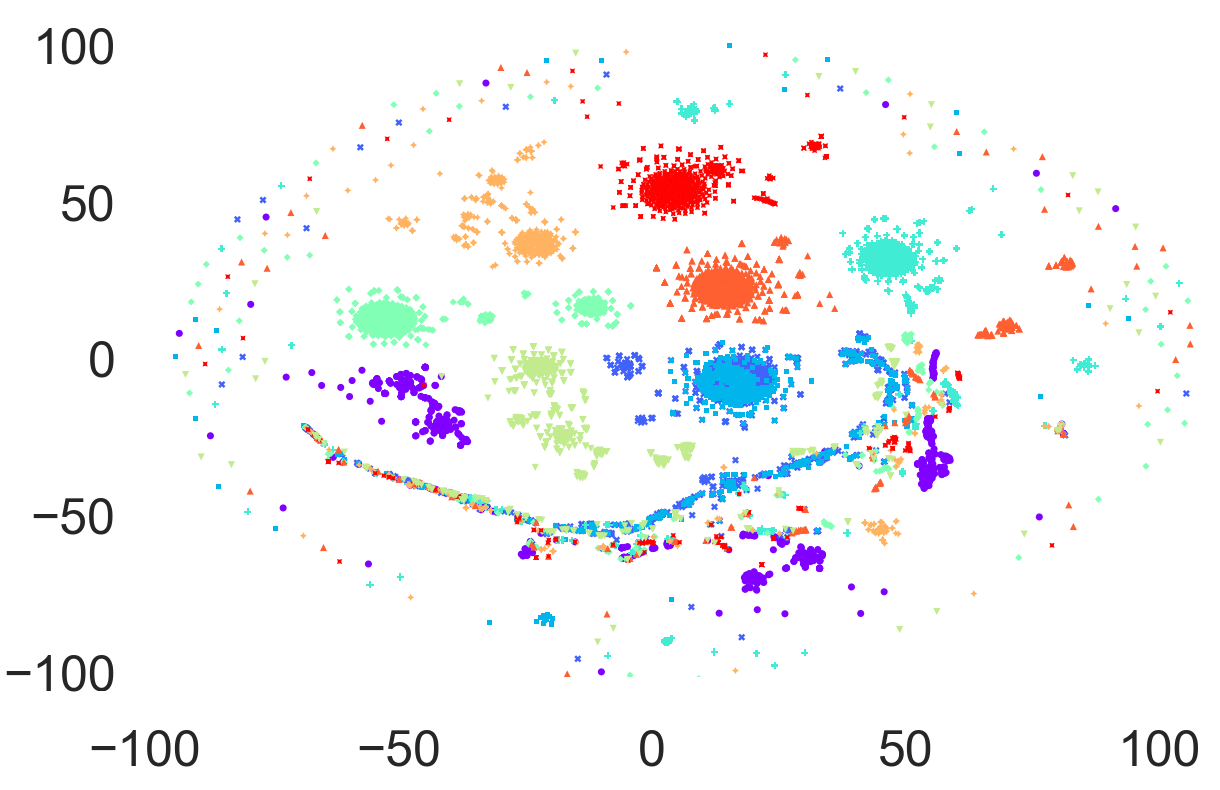}%
  \caption{String Kernel}
  \end{subfigure}%
  \begin{subfigure}{.25\textwidth}
  \includegraphics[scale=0.09]{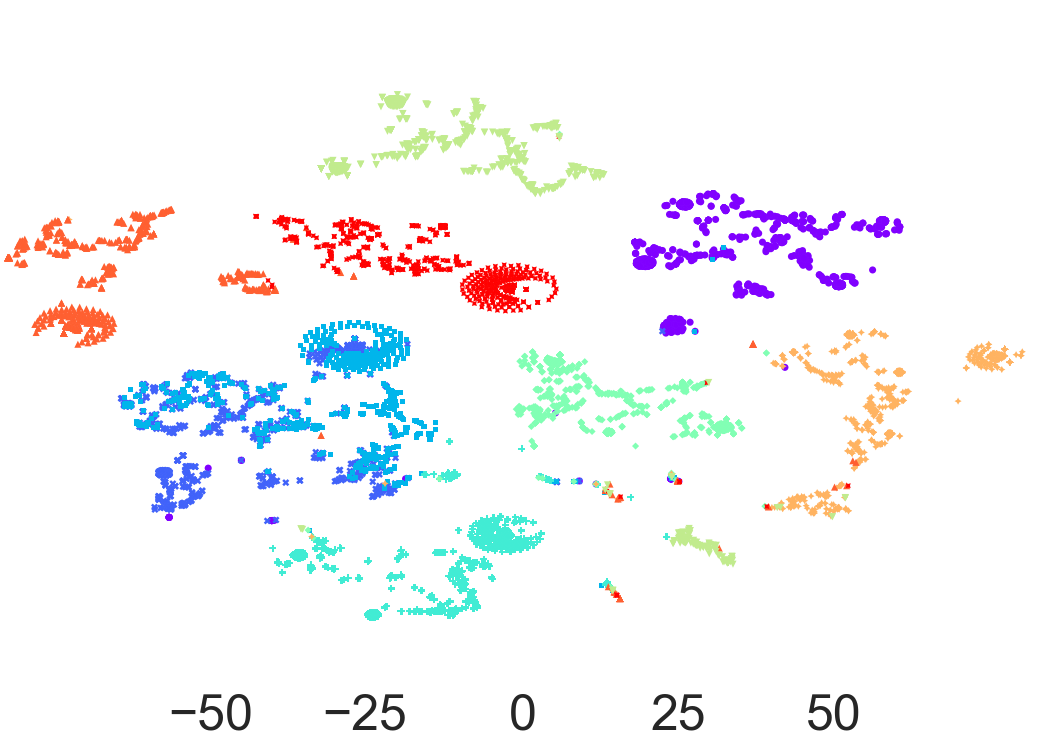}%
  \caption{WDGRL}
  \end{subfigure}%
  \\
  \begin{subfigure}{1\textwidth}
  \centering
  \includegraphics[scale=0.35]{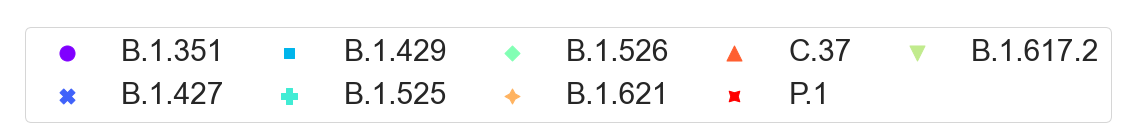}%
  \end{subfigure}%
  \caption{t-SNE plots for different embedding. The figure is best seen in color.
}
\label{fig_all_tsne}
\end{figure}

The t-distributed stochastic neighbor embedding (t-SNE)~\cite{van2008visualizing} is utilized to identify any hidden patterns in the data. This method works by mapping the high dimensional input data into $2D$ space but preserves the pairwise distance between data points. This visualization aims to highlight if different embedding methods introduce any changes to the overall distribution of the data. For various (baseline) embedding methods, Figure~\ref{fig_all_tsne} illustrates the t-SNE-based visualization (with SARS CoV-2 variants as labels shown in the legends). 
In the case of WDGRL, we can observe that the variants are not clearly grouped together. For Spike2Vec, PWM2Vec, and String Kernel, the majority of the variants, such as P.1 (Gamma), B.1.526 (Iota), and C.37 (Lambda), make a single group.

\section{Results and Discussion}\label{sec_results_discussion}
This section reports the classification results of the various methods using different evaluation metrics. We report the average and standard deviation scores of $5$ runs to minimize the effects of any randomness.

Table~\ref{tble_classification_results_variants} summarizes the results for our proposed system and the state-of-the-art (SOTA) models for different ML classifiers. We can observe that our proposed method with the LR classifier setting outperforms the baselines for all the evaluation metrics except for the training run time. 
While our method involves training multiple models which incurs high run times, it is able to preserve the privacy of data while maintaining the highest predictive performance, 
which is the prime goal of this paper. 
The federated learning-based model illustrates better performance than the feature-engineering-based baselines (Spike2Vec, PWM2Vec), achieving $3.3$\% and $0.4$\% more accuracy than the PWM2Vec and Spike2Vec methods, respectively, for the LR classifier. Similarly, it outperforms String Kernel with $0.4$\% accuracy using the LR classifier. Moreover, the proposed model outperforms WDGRL by $2.2\%$ and pre-trained Protein Bert by $2.9\%$ in terms of predictive accuracy using LR.

\begin{table*}[h!]
    \centering
    \resizebox{0.99\textwidth}{!}{
    \begin{tabular}{cp{1cm}cccccc|c}
    \toprule
        Method & Algo. & Acc. $\uparrow$ & Prec. $\uparrow$ & Recall $\uparrow$ & F1 (Weig.) $\uparrow$ & F1 (Macro) $\uparrow$ & ROC AUC $\uparrow$ & Train Time (Sec.) $\downarrow$\\
        \midrule \midrule
\multirow{7}{2cm}{Spike2Vec~\cite{ali2021spike2vec}}
 & SVM & 0.925 $\pm$ 0.001 & 0.926 $\pm$ 0.001 & 0.925 $\pm$ 0.001 & 0.924 $\pm$ 0.001 & 0.924 $\pm$ 0.002 & 0.958 $\pm$ 0.001 & 242.499 $\pm$ 4.623  \\
 & NB  & 0.919 $\pm$ 0.001 & 0.925 $\pm$ 0.003 & 0.919 $\pm$ 0.001 & 0.918 $\pm$ 0.001 & 0.918 $\pm$ 0.002 & 0.955 $\pm$ 0.001 & 6.452 $\pm$ 0.334 \\
 & MLP & 0.890 $\pm$ 0.015 & 0.894 $\pm$ 0.012 & 0.890 $\pm$ 0.015 & 0.889 $\pm$ 0.014 & 0.889 $\pm$ 0.013 & 0.938 $\pm$ 0.008 & 156.453 $\pm$ 14.703  \\
 & KNN & 0.866 $\pm$ 0.002 & 0.871 $\pm$ 0.002 & 0.866 $\pm$ 0.002 & 0.867 $\pm$ 0.002 & 0.866 $\pm$ 0.004 & 0.925 $\pm$ 0.002 & 16.039 $\pm$ 1.079  \\
 & RF  & 0.926 $\pm$ 0.003 & 0.927 $\pm$ 0.004 & 0.926 $\pm$ 0.003 & 0.925 $\pm$ 0.003 & 0.925 $\pm$ 0.003 & 0.958 $\pm$ 0.002 & 11.032 $\pm$ 0.175  \\
 & LR  & 0.927 $\pm$ 0.001 & 0.929 $\pm$ 0.002 & 0.927 $\pm$ 0.001 & 0.927 $\pm$ 0.001 & 0.927 $\pm$ 0.002 & 0.959 $\pm$ 0.001 & 23.966 $\pm$ 0.866  \\
 & DT  & 0.922 $\pm$ 0.004 & 0.924 $\pm$ 0.004 & 0.922 $\pm$ 0.004 & 0.922 $\pm$ 0.003 & 0.922 $\pm$ 0.002 & 0.956 $\pm$ 0.001 & 4.414 $\pm$ 0.172  \\
 \cmidrule{2-9}
 \multirow{7}{2cm}{PWM2Vec~\cite{ali2022pwm2vec}}
 & SVM & 0.888 $\pm$ 0.001 & 0.891 $\pm$ 0.001 & 0.888 $\pm$ 0.001 & 0.887 $\pm$ 0.002 & 0.885 $\pm$ 0.002 & 0.936 $\pm$ 0.001 & 13.718 $\pm$ 1.894 \\
 & NB  & 0.423 $\pm$ 0.014 & 0.449 $\pm$ 0.026 & 0.423 $\pm$ 0.014 & 0.352 $\pm$ 0.019 & 0.351 $\pm$ 0.017 & 0.675 $\pm$ 0.007 & 0.496 $\pm$ 0.047 \\
 & MLP & 0.866 $\pm$ 0.006 & 0.869 $\pm$ 0.008 & 0.866 $\pm$ 0.006 & 0.864 $\pm$ 0.006 & 0.862 $\pm$ 0.006 & 0.923 $\pm$ 0.003 & 12.656 $\pm$ 3.516 \\
 & KNN & 0.841 $\pm$ 0.010 & 0.843 $\pm$ 0.009 & 0.841 $\pm$ 0.010 & 0.841 $\pm$ 0.010 & 0.839 $\pm$ 0.009 & 0.910 $\pm$ 0.005 & 1.442 $\pm$ 0.181 \\
 & RF  & 0.899 $\pm$ 0.003 & 0.900 $\pm$ 0.003 & 0.899 $\pm$ 0.003 & 0.899 $\pm$ 0.003 & 0.897 $\pm$ 0.003 & 0.942 $\pm$ 0.002 & 6.608 $\pm$ 0.056 \\
 & LR  & 0.898 $\pm$ 0.004 & 0.898 $\pm$ 0.004 & 0.898 $\pm$ 0.004 & 0.896 $\pm$ 0.004 & 0.894 $\pm$ 0.004 & 0.941 $\pm$ 0.002 & 152.62 $\pm$ 7.102 \\
 & DT  & 0.882 $\pm$ 0.005 & 0.883 $\pm$ 0.005 & 0.882 $\pm$ 0.005 & 0.882 $\pm$ 0.005 & 0.880 $\pm$ 0.005 & 0.933 $\pm$ 0.003 & 3.406 $\pm$ 0.110 \\
 \cmidrule{2-9}
 \multirow{7}{2cm}{String Kernel~\cite{farhan2017efficient}}
 & SVM & 0.926 $\pm$ 0.005 & 0.931 $\pm$ 0.005 & 0.926 $\pm$ 0.005 & 0.924 $\pm$ 0.005 & 0.924 $\pm$ 0.003 & 0.959 $\pm$ 0.002 & 12.46 $\pm$ 2.543 \\
 & NB & 0.600 $\pm$ 0.008 & 0.705 $\pm$ 0.010 & 0.600 $\pm$ 0.008 & 0.611 $\pm$ 0.008 & 0.611 $\pm$ 0.008 & 0.775 $\pm$ 0.004 & 0.218 $\pm$ 0.013 \\
 & MLP & 0.853 $\pm$ 0.013 & 0.855 $\pm$ 0.014 & 0.853 $\pm$ 0.013 & 0.852 $\pm$ 0.013 & 0.853 $\pm$ 0.013 & 0.917 $\pm$ 0.007 & 6.948 $\pm$ 0.622 \\
 & KNN & 0.866 $\pm$ 0.007 & 0.872 $\pm$ 0.008 & 0.866 $\pm$ 0.007 & 0.868 $\pm$ 0.008 & 0.868 $\pm$ 0.005 & 0.925 $\pm$ 0.003 & 0.827 $\pm$ 0.068 \\
 & RF & 0.918 $\pm$ 0.004 & 0.919 $\pm$ 0.003 & 0.918 $\pm$ 0.004 & 0.917 $\pm$ 0.004 & 0.917 $\pm$ 0.002 & 0.954 $\pm$ 0.001 & 5.120 $\pm$ 0.191 \\
 & LR & 0.927 $\pm$ 0.004 & 0.930 $\pm$ 0.003 & 0.927 $\pm$ 0.004 & 0.926 $\pm$ 0.004 & 0.926 $\pm$ 0.002 & 0.959 $\pm$ 0.001 & 9.258 $\pm$ 0.702 \\
 & DT & 0.897 $\pm$ 0.006 & 0.899 $\pm$ 0.005 & 0.897 $\pm$ 0.006 & 0.897 $\pm$ 0.006 & 0.897 $\pm$ 0.004 & 0.942 $\pm$ 0.002 & 1.426 $\pm$ 0.065 \\
 \cmidrule{2-9}
 \multirow{7}{2cm}{WDGRL~\cite{shen2018wasserstein}}
 & SVM & 0.902 $\pm$ 0.003 & 0.905 $\pm$ 0.004 & 0.902 $\pm$ 0.003 & 0.901 $\pm$ 0.004 & 0.902 $\pm$ 0.003 & 0.946 $\pm$ 0.002 & 0.403 $\pm$ 0.038 \\
 & NB  & 0.825 $\pm$ 0.004 & 0.789 $\pm$ 0.007 & 0.825 $\pm$ 0.004 & 0.792 $\pm$ 0.004 & 0.795  $\pm$ 0.004 & 0.904 $\pm$ 0.002 & \textbf{0.016} $\pm$ 0.003 \\
 & MLP & 0.908 $\pm$ 0.004 & 0.910 $\pm$ 0.004 & 0.908 $\pm$ 0.004 & 0.907 $\pm$ 0.005 & 0.908 $\pm$ 0.004 & 0.949 $\pm$ 0.002 & 4.691 $\pm$ 0.736 \\
 & KNN & 0.910 $\pm$ 0.012 & 0.913 $\pm$ 0.011 & 0.910 $\pm$ 0.012 & 0.909 $\pm$ 0.012 & 0.910 $\pm$ 0.011 & 0.950 $\pm$ 0.006 & 0.116 $\pm$ 0.014 \\
 & RF  & 0.909 $\pm$ 0.002 & 0.911 $\pm$ 0.001 & 0.909 $\pm$ 0.002 & 0.907 $\pm$ 0.002 & 0.909  $\pm$ 0.002 & 0.949 $\pm$ 0.001 & 0.446 $\pm$ 0.057 \\
 & LR  & 0.877 $\pm$ 0.012 & 0.880 $\pm$ 0.005 & 0.877 $\pm$ 0.012 & 0.877 $\pm$ 0.015 & 0.878  $\pm$ 0.014 & 0.931 $\pm$ 0.006 & 0.168 $\pm$ 0.016 \\
 & DT  & 0.898 $\pm$ 0.005 & 0.900 $\pm$ 0.006 & 0.898 $\pm$ 0.005 & 0.897 $\pm$ 0.005 & 0.899  $\pm$ 0.004 & 0.943 $\pm$ 0.002 & 0.020 $\pm$ 0.005 \\
 \cmidrule{2-9}
 \multirow{4}{2cm}{Protein Bert~\cite{BrandesProteinBERT2022}}
 & \multirow{4}{*}{-} & \multirow{4}{*}{0.902 $\pm$ 0.004} & \multirow{4}{*}{0.903 $\pm$ 0.003} & \multirow{4}{*}{0.902 $\pm$ 0.004} & \multirow{4}{*}{0.904 $\pm$ 0.005} & \multirow{4}{*}{0.903 $\pm$ 0.009} & \multirow{4}{*}{0.945 $\pm$ 0.007} & \multirow{4}{*}{16127.76 $\pm$ 0.019} \\ 
 &&&&&&&&\\
 &&&&&&&&\\
 &&&&&&&&\\
 \cmidrule{2-9}
 \multirow{3}{2cm}{Federated Learning (ours)}
 & XGB  & 0.930 $\pm$ 0.004 & 0.932 $\pm$ 0.003 & 0.930 $\pm$ 0.004 & 0.930 $\pm$ 0.005 & 0.928 $\pm$ 0.004 & 0.960  $\pm$ 0.003 & 1578.27 $\pm$ 0.045 \\
 & LR  & \textbf{0.931} $\pm$ 0.011 & \textbf{0.933} $\pm$ 0.010 & \textbf{0.931} $\pm$ 0.012 & \textbf{0.931} $\pm$ 0.011 & \textbf{0.929} $\pm$ 0.011 & \textbf{0.961} $\pm$ 0.010 & 396.296 $\pm$ 0.024 \\
 & RF & 0.929 $\pm$ 0.005 & 0.932 $\pm$ 0.004 & 0.928 $\pm$ 0.006 & 0.927 $\pm$ 0.005 & 0.925 $\pm$ 0.006 & 0.959 $\pm$ 0.004 & 125.322 $\pm$ 0.079 \\
         \bottomrule
         \end{tabular}
    }
    \caption{Variants classification results (average $\pm$ standard deviation of 5 runs) for spike sequences data. The best average values are shown in bold.}
    \label{tble_classification_results_variants}
\end{table*}

The confusion matrix for the FL-based model using RF is shown in Table~\ref{tbl_confuse_mat_rf}. 
Similarly, the confusion matrix for the FL-based model using LR is shown in Table~\ref{tbl_confuse_mat_lr}.
We can observe that in most cases, the model is able to classify the variants correctly. An interesting observation here is in the results of variants B.1.427 and B.1.429. Since both of these variants are classified as Epsilon originating in California (see Table~\ref{tbl_variant_information}), the proposed model cannot distinguish between them because of their high similarity. Note that both of these variants share the same mutations in the spike region but have different mutations in other SARS-CoV-2 genes. Since we are dealing with spike regions in this study, differentiating between them becomes very difficult, that's why the model is getting confused between these two variants of Epsilon.

\begin{table}[h!]
\centering
    \begin{tabular}{cccccccccc}
    \toprule
          & B.1.351 & B.1.427 & B.1.429 & B.1.525 & B.1.526 & B.1.617.2 & B.1.621 & C.37 & P.1 \\
         \midrule
        B.1.351 & 283 & 0 & 0 & 1 & 4 & 3 & 0 & 0 & 0 \\
        B.1.427 & 0 & 173 & 140 & 0 & 4 & 0 & 0 & 0 & 0 \\
        B.1.429 & 1 & 48 & 267 & 0 & 1 & 0 & 0 & 1 & 1 \\
        B.1.525 & 1 & 1 & 0 & 287 & 1 & 0 & 0 & 0 & 0 \\
        B.1.526 & 0 & 0 & 0 & 1 & 297 & 0 & 0 & 0 & 0 \\
        B.1.617.2 & 0 & 0 & 0 & 0 & 0 & 283 & 0 & 0 & 0 \\
        B.1.621 & 0 & 0 & 0 & 0 & 2 & 0 & 296 & 0 & 0 \\
        C.37 & 1 & 0 & 1 & 0 & 1 & 0 & 0 & 297 & 0 \\
        P.1 & 0 & 0 & 0 & 0 & 0 & 0 & 0 & 0 & 304 \\
        \bottomrule
    \end{tabular}
     \caption{Random Forest}
    \label{tbl_confuse_mat_rf}
\end{table}

\begin{table}[h!]
\centering
    \begin{tabular}{cccccccccc}
    \toprule
          & B.1.351 & B.1.427 & B.1.429 & B.1.525 & B.1.526 & B.1.617.2 & B.1.621 & C.37 & P.1 \\
         \midrule
        B.1.351 & 302 & 0 & 0 & 0 & 0 & 0 & 0 & 0 & 0 \\
        B.1.427 & 0 & 166 & 138 & 0 & 1 & 0 & 0 & 0 & 0 \\
        B.1.429 & 1 & 57 & 262 & 1 & 0 & 0 & 0 & 0 & 0 \\
        B.1.525 & 0 & 0 & 1 & 285 & 0 & 3 & 0 & 0 & 0 \\
        B.1.526 & 0 & 1 & 0 & 0 & 309 & 0 & 0 & 0 & 0 \\
        B.1.617.2 & 0 & 0 & 0 & 0 & 0 & 293 & 0 & 0 & 0 \\
        B.1.621 & 0 & 0 & 0 & 0 & 1 & 0 & 297 & 0 & 0 \\
        C.37 & 0 & 0 & 0 & 0 & 0 & 0 & 0 & 306 & 0 \\
        P.1 & 1 & 0 & 2 & 0 & 0 & 0 & 0 & 0 & 273 \\
        \bottomrule
    \end{tabular}
  \caption{Logistic Regression}
 \label{tbl_confuse_mat_lr}
\end{table}

\subsection{Local Model Analysis}
\begin{figure}[h!]
  \centering
  \begin{subfigure}{.33\textwidth}
  \centering
  \includegraphics[scale=0.50]{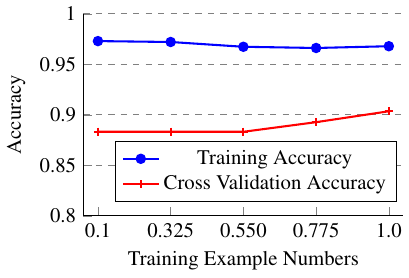}%
  \caption{Local Model 1}
  \end{subfigure}%
  \begin{subfigure}{.33\textwidth}
  \includegraphics[scale=0.50]{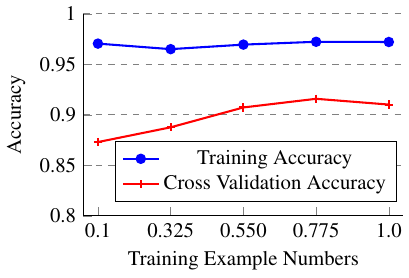}%
  \caption{Local Model 2}
  \end{subfigure}%
  \begin{subfigure}{.33\textwidth}
  \centering
  \includegraphics[scale=0.50]{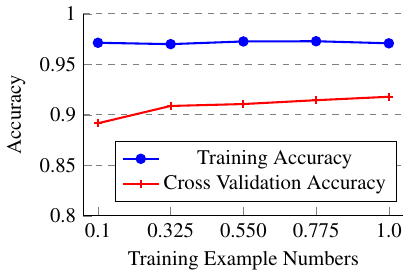}%
  \caption{Local Model 3}
  \end{subfigure}%
  \caption{Training and Cross-Validation accuracy of different local models with increasing (fraction of) training set size (x-axis). The figure is best seen in color.}
\label{fig_local_model_results}
\end{figure}

We present the training and validation accuracy for individual ML models in Figure~\ref{fig_local_model_results} to assess the performance of individual models throughout the training phase. We can observe that these charts demonstrate accuracy improvements as the training set size increases, showing the improvement of the model.

\subsection{Global Model Analysis}
\begin{figure}[h!]
\centering
\begin{subfigure}{.5\textwidth}
  \centering
  \includegraphics[scale=0.8]{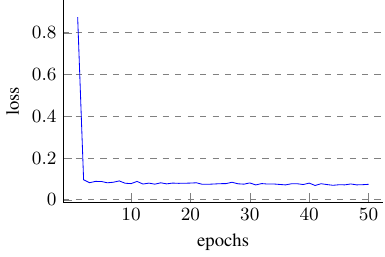}
  \caption{Loss}
  \label{fig_loss_NN}
\end{subfigure}%
\begin{subfigure}{.5\textwidth}
  \centering
  \includegraphics[scale=0.8]{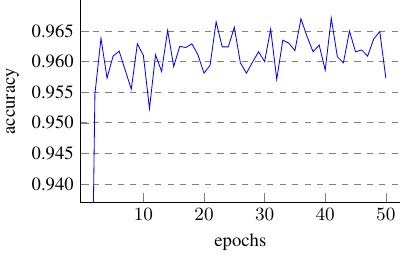}
  \caption{Accuracy}
  \label{fig_acc_NN}
\end{subfigure}
\caption{Loss and Accuracy of final ensemble (Global) model (NN).}
\label{fig_model_loss_accuracy}
\end{figure}

The accuracy and loss curves for the global model are shown in Figure~\ref{fig_model_loss_accuracy}. We can observe in Figure~\ref{fig_loss_NN} that the loss is stable after $20$ epochs, and accuracy ranges around 94-96\% as shown in Figure~\ref{fig_acc_NN}.

\subsection{Laws of Federated Learning}
In this section, we discuss the different laws of federated learning that the proposed model holds.

\paragraph{Law of data ownership}
This law is upheld in the proposed model since the data is kept locally on each local model's device, and only the model outputs are shared among the devices.

\paragraph{Law of data privacy}
This law is upheld in the proposed algorithm since the data is not shared between the devices, only the model parameters are shared.

\paragraph{Law of model aggregation}
This law is upheld in our model since the model parameters from each participant are combined at a central server to create a global model.

\paragraph{Law of model heterogeneity}
This law is upheld in our algorithm since each participant may use a different local training algorithm and hyperparameters to train their model.

\section{Conclusion}\label{sec_conclusion}
We propose federated learning-based models for SARS-CoV-2 variant classification. We show that by using spike sequences only, we can achieve good predictive performance. We compare the results using different evaluation metrics with several SOTA models and show that the federated learning-based approach outperforms those existing models from the literature. An important property of the proposed model is that since it only transfers the output from local models to the global model, it preserves the privacy of users, which could be a major problem in many big organizations. Especially in healthcare addressing the issue of privacy is of major concern and the proposed model addresses the issue while not compromising the performance.
One possible extension of this approach is to apply deep learning-based local models to classify the sequences. Another interesting direction would be to propose an approximate approach to compute feature embeddings for the biological sequences to further improve computational overhead. Using different ML classifiers in combination within a single FL architecture could also be an interesting future extension for SARS-CoV-2 variant classification. We will also explore incorporating other attributes (e.g., regions, time) and variants along with the spike sequences to generate a vertical federated learning model.
Investigating the generalization of the proposed model to other protein region sequences is also an exciting future direction.

\bibliographystyle{splncs04}
\bibliography{25}

\end{document}